\documentclass[conference]{IEEEtran}
\IEEEoverridecommandlockouts
\usepackage{cite}
\usepackage{amsmath,amssymb,amsfonts}
\usepackage{algorithmic}
\usepackage{graphicx}
\usepackage{textcomp}
\usepackage{xcolor}
\usepackage{comment}
\usepackage{siunitx}
\usepackage{tikz}
\usepackage{pgfplots}
\pgfplotsset{compat=1.5}

\newcommand{\plotFont}{\small}

\newcommand\copyrighttext{%
  \footnotesize \textcopyright This work has been submitted to the IEEE for possible publication. Copyright may be transferred without notice, after which this version may no longer be accessible.}
\newcommand\copyrightnotice{%
\begin{tikzpicture}[remember picture,overlay]
\node[anchor=south,yshift=10pt] at (current page.south) {\fbox{\parbox{\dimexpr\textwidth-\fboxsep-\fboxrule\relax}{\copyrighttext}}};
\end{tikzpicture}%
}

\begin{document}

\title{Beyond Task Performance: \\ Human Experience in Human-Robot Collaboration\\
\thanks{$^{1}$The authors are with the Institute of Control Systems (IRS), Karlsruhe Institute of Technology (KIT), Kaiserstr. 12, Karlsruhe 76131, Germany {\tt\small sean.kille@kit.edu}} %
\thanks{$^{2}$The authors are with the Institute of Product Engineering (IPEK), Karlsruhe Institute of Technology (KIT), Kaiserstr. 12, Karlsruhe 76131, Germany}%
\thanks{$^{3}$The authors are with Autonomous Learning Robots, Institute of Antropomatics and Robotics (IAR), Karlsruhe Institute of Technology (KIT), Kaiserstr. 12, Karlsruhe 76131, Germany}%
\thanks{$^{4}$The authors are with the Cognition, Action, and Sustainability Unit, Department of Psychology, University of Freiburg, Engelbergerstr. 41, Freiburg 79085, Germany  }%
\thanks{This work involved human subjects in its research. Approval of all ethical and protocols was granted by the Ethics Committee of KIT.}
\thanks{This research was funded by the Federal Ministry of Education and Research (BMBF) and the Baden-Württemberg Ministry of Science as part of the Excellence Strategy of the German Federal and State Governments. This work was also funded by the Deutsche Forschungsgemeinschaft (DFG, German Research Foundation).}
}

\author{Sean Kille$^{1}$, Jan Heinrich Robens$^{2}$, Philipp Dahlinger$^{3}$, Alejandra Rodríguez-Velásquez$^{4}$, Simon Rothfuß$^{1}$, \\ Balint Varga$^{1}$, Andreas Lindenmann$^{2}$, Gerhard Neumann$^{3}$, Sven Matthiesen$^{2}$, Andrea Kiesel$^{4}$, Sören Hohmann$^{1}$
}

\maketitle

\copyrightnotice

\begin{abstract}
Human interaction experience plays a crucial role in the effectiveness of human-machine collaboration, especially as interactions in future systems progress towards tighter physical and functional integration. While automation design has been shown to impact task performance, its influence on human experience metrics such as flow, sense of agency (SoA), and embodiment remains underexplored. This study investigates how variations in automation design affect these psychological experience measures and examines correlations between subjective experience and physiological indicators. A user study was conducted in a simulated wood workshop, where participants collaborated with a lightweight robot under four automation levels. The results of the study indicate that medium automation levels enhance flow, SoA and embodiment, striking a balance between support and user autonomy. In contrast, higher automation, despite optimizing task performance, diminishes perceived flow and agency. Furthermore, we observed that grip force might be considered as a real-time proxy of SoA, while correlations with heart rate variability were inconclusive. The findings underscore the necessity for automation strategies that integrate human-centric metrics, aiming to optimize both performance and user experience in collaborative robotic systems. 
\end{abstract}


\section{Introduction} \label{sec:intro}

Over the past decades, advancements in human-machine interaction (HMI) have led to a more intertwined relationship between humans and machines.  
The development of HMI can be categorized into several (partly overlapping) stages~\cite{Schmidtler.2015, Matheson.2019}. Initially, human-machine coexistence was characterized by shared workspaces without the pursuit of a joint task.
The next level is human-machine cooperation, where a common task requires communication between the agents. 
When the interaction involves close physical contact, cooperation evolves into human-machine collaboration. 
This stage is characterized by haptic communication, including close physical coupling for direct interaction, with the machine often being a robot.

Ergonomic design plays a key role in shaping performance and human experience in HMI. 
On one hand, ergonomic design affects objective measures, such as the impact of tool handle design on performance and injury prevention~\cite{Harih.2013, Fellows.1991}.
On the other hand, ergonomic design also shapes subjective experiences, reflected in measures like comfort~\cite{Schiele.2006, Uhl.2021} or perceived suitability of usage of tools~\cite{Germann.2019}.

Beyond static hardware design, the automation design of machines also plays a crucial role in shaping performance and human experience~\cite{Abbink.2012}. Automation design strongly influences task performance, which is often the primary objective, e.g., in surgical HMI applications~\cite{Du.2024}. 
But as interactions become more closely coupled, the impact of automation design on human interaction experience increases, potentially leading to a new stage in HMI termed human-machine symbiosis~\cite{Inga.2022}. 
Currently however, the relationship between the machine's automation design and human experience is not well-explored in the literature. Some measures analyzing the joint interaction have gained increased interest, such as trust~\cite{Shi.2022}, or have been newly introduced, e.g., fluency~\cite{Hoffman.2019a}.
However, other subjective experience measures such as flow~\cite{Csikszentmihalyi.1989}, sense of agency (SoA)~\cite{Haggard.2012}, and embodiment~\cite{Longo.2008} become particularly relevant in prolonged interactions but remain insufficiently studied in HMI contexts.

Flow, as described by \cite{Csikszentmihalyi.1989}, enhances task engagement and efficiency, making it essential in automated environments where sustained attention is required~\cite{Hancock.2019}. In human–machine collaboration, the achievement of flow depends on the balance of task challenges with user skills. Research shows that adjusting task difficulty in industrial assembly scenarios can facilitate flow, leading to improved performance and user satisfaction~\cite{Prajod.2024}. Similarly, maintaining a strong SoA ensures that users feel in control of automated processes, reducing over-reliance or automation complacency~\cite{Wen.2015}. In human–robot collaboration, a heightened sense of embodiment can improve task performance and user acceptance of robotic assistance. For instance, inducing embodiment through multisensory bodily interactions can increase a robot’s acceptability, positively influencing social attitudes toward the machine~\cite{Ventre-Dominey.2019}. However, it remains unclear if and how these experience measures can be influenced by automation design in an HMI.

This paper aims to fill this gap by experimentally investigating how automation design influences key human experience measures like flow, SoA, and embodiment. Insights in this field, together with advancements in precisely and quickly identifying human behavior, e.g. as shown in~\cite{Karg.2024}, pave the way for adaptive human-machine cooperation~\cite{Varga.2024} and human-centered control design~\cite{Kille.2024a}.

While ergonomic hardware design is fixed during product development, a machine’s automation can be continuously adapted, even during interaction. 
Currently, human experience is often assessed through self-report questionnaires, which are conducted post-interaction or during the interaction, thus not allowing for real-time measurement and potentially disrupting the interaction~\cite{Moneta.2012, Jackson.1996, Engeser.2008}.  
Recent research explores real-time assessment of human experience using physiological measures such as heart rate~\cite{ Knierim.2018, Harmat.2015} , heart-rate variability (HRV)~\cite{Peifer.2014,Rissler.2023}, electro-dermal activity~\cite{Klarkowski.2016}, and EEG~\cite{Labonte-LeMoyne.2016}. We aim to advance this field by analyzing the connection between physical measures and psychological experience, and examining the reproducibility of correlations between HRV and flow in human-machine interactions.

In the paper, we make two key contributions: 
first, we analyze how the automation design of a machine affects human psychological experience during interaction. 
Second, we investigate the relationship between human experience and physical and physiological measures. The experimental design and automation developed for this investigation are presented in Section~\ref{sec:studyDesign}. The obtained results are laid out in Section~\ref{sec:results}, followed by their discussion in Section~\ref{sec:discussion}. The paper concludes with Section~\ref{sec:conclusion}.

\section{Study Design} \label{sec:studyDesign}

In this section, we present the design of our study, which investigates how  variations in automation design impact human experience. 

\subsection{Study task} \label{sec:task}
\begin{figure}[t]
    \centerline{\includegraphics[width=0.48\textwidth,keepaspectratio]{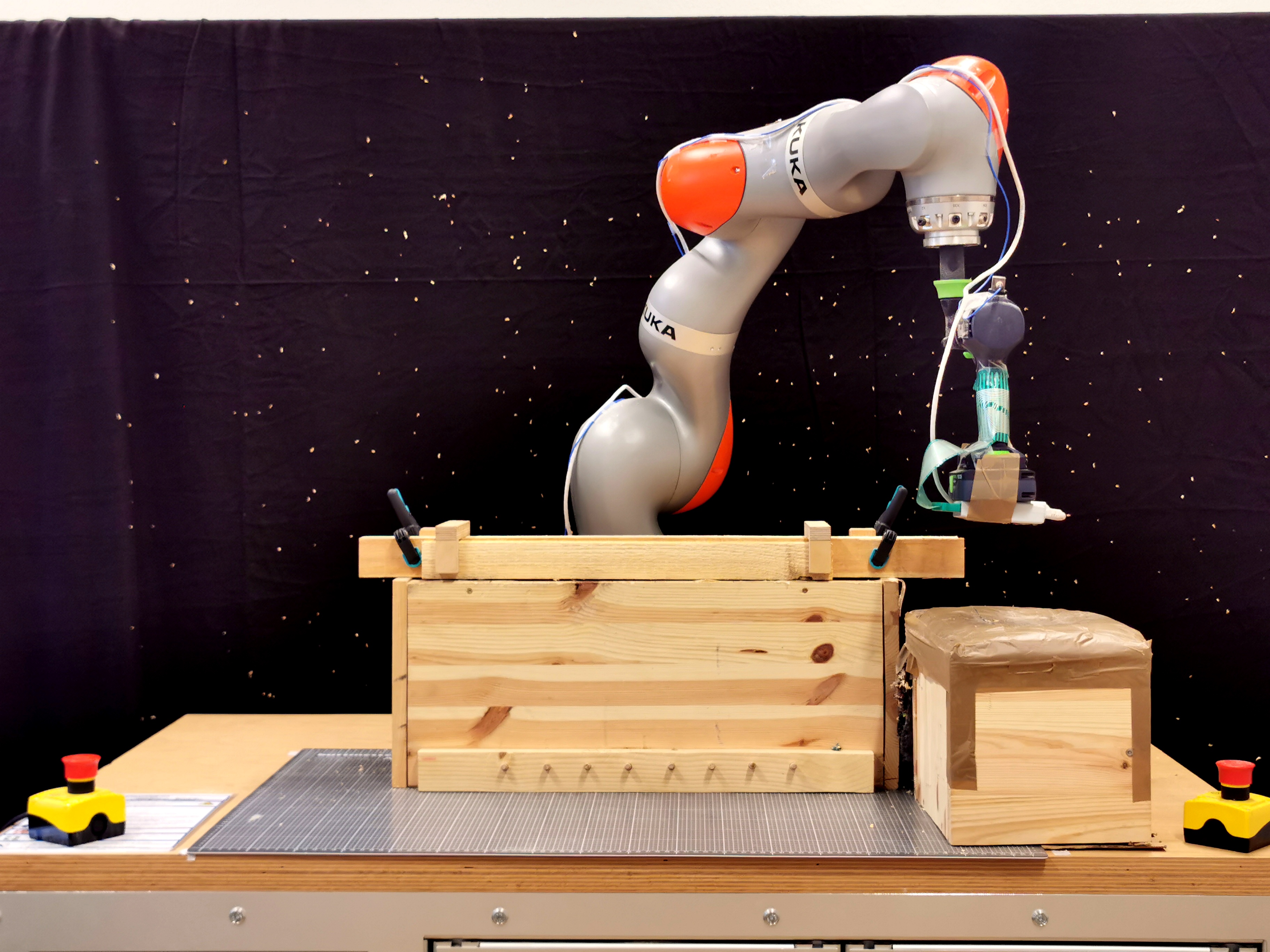}} %
    \caption{Cordless drill mounted to a KUKA LBR iiwa R820 on a workshop workbench. Positions and interaction forces are measured using the robot, thin film pressure sensors measure grip forces.} 
    \label{fig:workbench}
\end{figure}

Our experimental design simulates a wood workshop where a human and a machine collaboratively perform woodwork using a cordless screwdriver.   
The human grasps the cordless drill conventionally, which is connected to a light-weight robot that assists in the shared task. 
Raw wooden boards are mounted to a stand in the center of the workspace, and the cordless drill rests on a box next to the stand by default (see Fig.~\ref{fig:workbench} for the complete setup).

The task involves drilling holes into wooden boards to create a coat rack.
Specifically, eight equidistant holes must be drilled on a horizontal level, inclined downwards by $-15^{\circ}$ from the perpendicular axis of the board. No other specifications are imposed. 

\subsection{Experimental hardware} 

The cordless drill used is a Festool TPC 18/4 (Festool GmbH, Wendlingen am Neckar, Germany) weighing $\SI{2.3}{\kilogram}$ including the battery pack, with an $\SI{8}{\mm}$ wood drill bit. 
A custom adapter attaches the drill to a lightweight KUKA LBR iiwa 14 R820 robot (KUKA Deutschland GmbH, Augsburg, Germany) (LBR) \cite{KUKADeutschlandGmbH.2019}.   
The LBR has seven axes, providing seven degrees of freedom (DOF), and operates within a spherical shell workspace with an outer diameter of \SI{820}{\mm} and an inner diameter of \SI{420}{\mm}. 
It can apply forces up to \SI{140}{\N} and torques up to \SI{40}{Nm}. 
The LBR is controlled by a KUKA Sunrise Workbench. 

This setup allows for real-time acquisition of pose and force data from the LBR to MATLAB and the communication of set-values from MATLAB to the LBR controller. 
We use an impedance controller similar to~\cite{Braun.2023}, which sets wrenches corresponding to a force $\boldsymbol{F}_{\text{i}}$. 
This force $\boldsymbol{F}_{\text{i}}$ depends on the measured cartesian LBR pose $ \boldsymbol{p}_{\text{m}}\in\mathbb{R}^{6}$, a  set-pose in cartesian space $ \boldsymbol{p}_{\text{s}}\in\mathbb{R}^{6}$ and a stiffness matrix $\boldsymbol{K}\in\mathbb{R}^{6\times6}$:     

\begin{align}
    \boldsymbol{F}_{\text{i}} = \boldsymbol{K}\left(\boldsymbol{p}_{\text{m}}-\boldsymbol{p}_{\text{s}} \right). \label{eq:imp}
\end{align}

The diagonal matrix $\boldsymbol{K}$ represents the stiffness of virtual springs for each of the six DOFs in cartesian space as the diagonal elements ($K_{ii}\geq 0, i\in \left\{1,\dots,6\right\}$). 
It is therefore possible to set individual forces for each DOF using the two parameters set-pose $p_{\text{s},i}$ and stiffness $K_{ii}$.

\subsection{Automation parametrization} \label{subsec:automationParam}
\begin{figure}[t]
    \centerline{\includegraphics[width=0.48\textwidth,keepaspectratio]{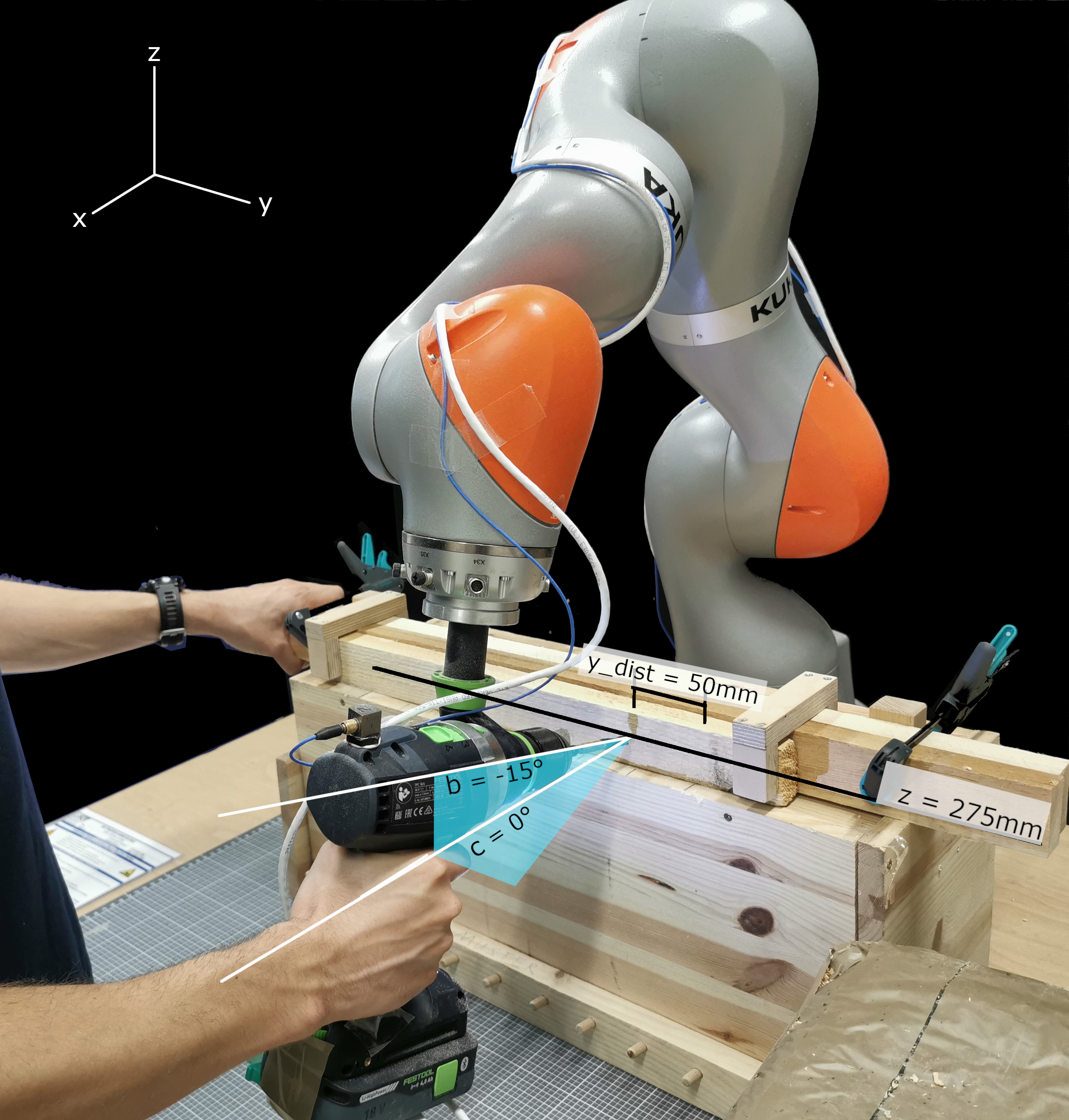}} %
    \caption{The human operates the cordless drill while being assisted by the robot. The task specifications in rotational and translational dimensions are highlighted.} 
    \label{fig:workbenchTask}
\end{figure}

The automation design aims to support the human in following the task specifications while ensuring that the human feels responsible for the overall action. The automation helps to find the specified drill positions without performing movements independently of the human. The drill setpoints in rotational dimensions are $b = \SI{-15}{\degree}$ downward pitch, no yaw ($c=\SI{0}{\degree}$), and no specification regarding the roll $a$. In the translational dimensions, the holes are to be placed equidistantly in the  y-direction (marked on the stand in $\SI{50}{\mm}$ intervals) and horizontally level in the center of the wooden board ($z=\SI{275}{\mm}$) (see  Fig.~\ref{fig:workbenchTask} for specified axes and positions).

The automation is implemented separately for rotational and translational spaces. For the rotational axes $a$, $b$ and $c$, the automation smoothly rotates the cordless drill towards the specified setpoints as it approaches the wooden board. This is realized through a linear increase in the rotational stiffness $K_{r}$  (with $r \in \left\{4,5,6\right\}$) depending on the distance between the measured $x$-position $x_{\text{m}}$ of the drill bit and the wooden board position $x_{\text{w}}$, starting from a threshold of $x_{\text{th,R}} = \SI{200}{\mm}$:  

\begin{align}
    K_{r} = 
\begin{cases}
    k_{r},                                            & \text{for } x_{\text{m}} < x_{\text{w}} \\
    \frac{ x_{\text{th,R}} - |x_{\text{}m}-x_{\text{w}}|}{x_{\text{th,R}}}  k_{r},     & \text{for } 0 \leq x_{\text{m}} - x_{\text{w}} < x_{\text{th,R}} \\
    0,                                                       & \text{for } x_{\text{w}} + x_{\text{th,R}} \leq x_{\text{m}}.
\end{cases}
\label{eq: stiffRot}
\end{align}

Here, $K_{r}$ represents the stiffness for all rotational axes $a$, $b$ and $c$. 
If the measured drill bit position $x_{\text{m}}$ is more than $x_{\text{th,R}}= \SI{200}{mm}$ from the wooden board, no rotational stiffness is applied, allowing free movement. 
Within the threshold space, stiffness increases linearly as the drill bit approaches the board. Upon reaching the board, the stiffness for all rotational dimensions $K_r$ is maintained at the maximum value $k_r$ to provide maximum support.

The translational support is implemented similarly. Stiffness depends on the translational distance between the closest drill axis $p_{\text{t,ax}}$ and the measured position $p_{\text{t,m}}$: $d_{\text{d}} = |p_{\text{t,ax}}-p_{\text{t,m}}|$. This distance $d_{\text{d}}$ is mapped onto a sigmoid function, creating a magnetic-like effect that pulls the drill bit tip towards the closest point on the nearest drill axis. The drill axes are defined up to a distance of $\SI{100}{\mm}$ from the board, so the magnetic effect is only present near the working area. For a distance $d_{\text{d}}$ greater than $x_{\text{th,T}}=\SI{30}{\mm}$ no translational stiffness is applied:

\begin{align*}
    K_{t} = 
\begin{cases}    
    k_{t} \biggl( 1+    \text{tanh} \Bigl(   
     \frac{(|x_{\text{th,T}} -  d_{\text{d}}|)-\frac{x_{\text{th,T}}}{2}}{2} \Bigr)  \biggr),    & \text{for } d_{\text{d}} < x_{\text{th,T}} \\    
    0,                                                                       & \text{for } x_{\text{th,T}} \leq d_{\text{d}}.  
\end{cases}
\label{eq: stiffTransl}
\end{align*}

The stiffness parameters for each of the four modes are shown in Table~\ref{table:stiffParam}. These modes are defined as follows: M0 provides no support for reaching the specified set positions and leaves most of the cordless drill's weight to be carried by the human. With the LBR attached to the power drill, the user experiences additional inertia from the robotic arm. To mitigate this, $\SI{0.5}{\kg}$ of the gravitational force is supported in this mode, resulting in a more pleasant behavior. In modes M1-M3, the full weight of the cordless drill plus all attached sensory equipment of $\SI{2.5}{\kg}$ is compensated, allowing the cordless drill to float when released by the human operator. The support increases from M1 to M3, with M1 offering subtle assistance and M3 providing firm stiffness with minimal error tolerance. 

\begin{table}[tbp]
    \caption{Automation Mode Parameters}
\begin{center}
    \begin{tabular}{|c | c | c | c | c|}
        \hline 
         & M0 & M1 & M2 & M3 \\
        \hline \hline
        
        $k_{\mathrm{a}}$ in \si{N\m}/\si{\radian} & 0 & 5 & 10 & 20   \\
        \hline
        $k_{\mathrm{b}}$ in \si{N\m}/\si{\radian}  & 0 & 10 & 30 & 50 \\
        \hline
        $k_{\mathrm{c}}$ in \si{N\m}/\si{\radian} & 0 & 10 & 20 & 25 \\
        \hline
        $x_{\text{th,R}}$ in \si{\mm} & \multicolumn{4}{|c|}{200} \\
        \hline \hline
        $k_{\mathrm{x}}$ in \si{\N}/\si{m} & 0 & 0 & 0 & 0 \\
        \hline
        $k_{\mathrm{y}}$ in \si{\N}/\si{m} & 0 & 40 & 150 & 210  \\
        \hline
        $k_{\mathrm{z}}$ in \si{\N}/\si{m} & 0 & 80 & 400 & 520 \\
        \hline
        $x_{\text{th,T}}$ in \si{\mm} & \multicolumn{4}{|c|}{30} \\
        \hline \hline
        $m_{\text{grav}}$ in \si{\kg} & $\num{0.5}$ & \multicolumn{3}{|c|}{\num{2.5}} \\
        \hline
    \end{tabular}
    \label{table:stiffParam}
\end{center}
\end{table}

\begin{figure*}[tbp]
    \centering
    \resizebox{\textwidth}{!}{
        \input{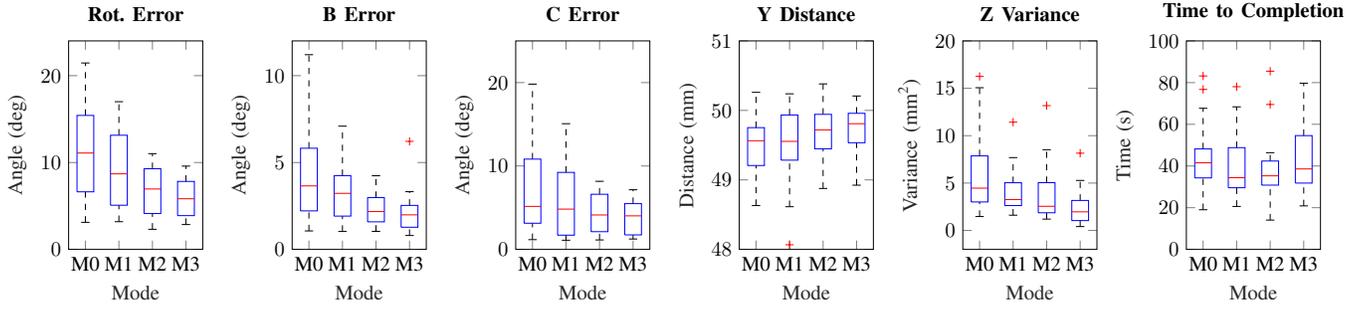}
        }
    \caption{Box plots showing rotational errors (Rot. Error being the sum of $B$ Error and $C$ Error), $y$~distance, $z$~variance, and time to completion for different support levels (M0–M3). Mean rotational errors and variance and $z$~variance decrease with increasing level of support. The mean $y$~distance approaches the task specification with higher level of support. No differences in time to completion can be observed. }
    \label{plot:boxObjective}
\end{figure*}

\subsection{Experimental procedure}
The following outlines the procedure that each participant followed in the experimental study, which takes approximately $\num{40}-\SI{45}{\min}$. 

\subsubsection{Introduction and preparation}
First, participants receive a general introduction to the experiment. We then explain how the data will be used and obtain their consent for participation and data usage. Participants are asked to position a heart rate monitor on their sternum. After verifying correct placement, participants are asked to watch a $\SI{5}{\min}$ video of a virtual aquarium in order to normalize their heart rate to acquire a resting heart rate and mentally prepare for the study. Finally, a $\SI{3}{\min}$ long video explains the procedure of the study and provides safety instructions. 

\subsubsection{Main part}

In this part of the study, each participant interacts with all four automation modes described in Section~\ref{subsec:automationParam}. The order of modes is counterbalanced through pseudo-randomization to account for learning effects. In each mode, participants drill eight holes into four wooden boards, following the specifications in Section~\ref{sec:task}. After each board is drilled, participants answer one verbal question. At the end of each mode, they complete a questionnaire assessing their experience. The study concludes with participants completing a demographic questionnaire.

\subsection{Data acquisition}

\subsubsection{Human experience} \label{subsubsec:humExp}
Human experience in our study is assessed using state-of-the-art questionnaires. After each mode, participants completed a 24-question survey. The first 13 questions comprise the Flow Short Scale (FSS) by Rheinberg \cite{Rheinberg.2003}, measuring \textit{flow} experience and subcategories such as \textit{smooth process}, \textit{absorption} and \textit{concern}. Three further questions assess the perceived difficulty. The measures \textit{SoA} and \textit{embodiment} are determined with three questions each, adapted from \cite{Tapal.2017, Gonzalez-Franco.2018} to fit our experiment design. The questionnaire concludes with questions about participant's satisfaction with their work result and estimated work duration. All questions, except the last, are answered on a 7- or 9-point Likert scale, depending on the question's origin. Additionally, after each board is complete, one question of the FSS is posed verbally and documented by the study tutor to provide detailed experience documentation without significantly distracting the participant. 

\subsubsection{Position and interaction force}
Since the human and the automation collaboratively work with the cordless drill, and the LBR is coupled permanently to the drill, the position of the drill bit tip can be tracked via the LBR. Additionally, the translational and rotational interaction forces between the human and the LBR are recorded. The LBR records the data with a sample time of~\SI{3.5}{\ms}. 

\subsubsection{Grip force} 
The grip force exerted by the human on the cordless drill handle  is captured using a thin film pressure sensor (F-Scan 3000E, Tekscan Inc., Norwood, MA, USA ; sample rate: \SI{100}{Hz}; 3.9 sensor points per $cm^2$) that wraps around the handle. 

\subsubsection{Heart rate sensor}
As a physiological measure, participants wear a Polar H10 heart rate chest strap device (Polar Electro Oy, Kempele, Finland; sampling rate: 1000 Hz). The device collects RR intervals and beats per minute (bpm) measurements.

\subsection{Analysis method}

\paragraph{Objective measures}
To objectively assess task performance, we evaluate how well the task specifications are met, given in~\ref{sec:task}. First, we calculate the mean drilling angle maintained while the drill bit is within the wooden board. The deviation from the specified angle in the b- and c-dimensions is then determined, and the sum of these deviations defines the rotational error. To assess compliance with the specified hole spacing of $\SI{50}{mm}$, we calculate the mean distance between the eight holes and average this value over the four boards per mode. The levelness of drilling height is evaluated by computing the variance in the $z$-dimension across all eight holes, which is then averaged over the four boards per mode. Additionally, completion time is recorded for each board and averaged across the four boards per mode.

\paragraph{Physiological responses}
To analyze physiological responses, we measure grip force by recording the participants' mean grip force on the cordless drill handle while the drill bit is positioned in front of the wooden boards. Regarding HRV, we compute the mean and standard deviation of heart rate and RR intervals, along with the root mean square of successive differences (RMSSD) of RR intervals.

\paragraph{Subjective measures}

Subjective experience is assessed using multiple self-reported measures. The \textit{flow} score is calculated based on the 13 questions of the FSS, with the overall score determined by averaging all responses after inverting the negatively phrased items. \textit{Embodiment} and \textit{SoA} are each determined by averaging the responses to their respective three-item scales. Additionally, work result satisfaction is assessed using the rating from the corresponding questionnaire item.

\paragraph{Correlation analysis}

For the analysis of correlations, We use a repeated-measures one-way ANOVA, applying a Greenhouse-Geisser (GG) correction if the sphericity assumption does not hold. For pairwise comparisons, we use a t-test with Bonferroni-correction. A difference is considered significant if the test provides a p-value of $p<0.05$.  
For the determination of effect size, we follow the effect size guidelines by~\cite{Cohen.1992}.

\section{Results} \label{sec:results}
In this section, we present the results gathered from our experimental study. A total of 28 participants took part in the within-groups experimental design, of which two had to be excluded due to faulty data logging. The remaining participants included 24 men and two women, 24 in their 20s, two in their 30s, and one in their 50s. All participants operated the drill with their right hand. We performed statistical analysis on various metrics collected during the experiment.

\subsection{Control accuracy} \label{sec:resultsObjective}

For objective performance analysis, we evaluated how well the task specifications were met under the four different levels of automation. 
The mean position and variance of the drill bit, while within the volume of the wooden board, were extracted for each hole and aggregated over all holes drilled in one mode. 

In terms of rotational error, we observed that with increasing support, the mean rotational error (considering the error in the rotational axis from $b=15^{\circ}$ and $c=0^{\circ}$) showed a significant negative correlation ($r=-0.44$ and $p<0.001$), indicating a medium effect.
For the translational dimensions, we analyzed the variance in height by calculating the $z$-variance per board and aggregating the mean over all boards per mode. 
A weak negative correlation ($r=-0.27$ and $p=0.005$) was found between the $z$-variance and the level of support. 
The variance in equidistance was determined through the variance of the $y$-distance between the drilled holes. 
No significant correlation was found between the levels of support and the mean variance in equidistance ($r=-0.012$ and $p=0.91$). 
Regarding the completion time, a repeated-measures one-way ANOVA showed no significant difference in the mean duration to complete each board across the modes ($p=0.35$): $t_{\text{M0}} = \SI{41.6}{\s}$, $t_{\text{M1}} = \SI{34.4}{\s}$, $t_{\text{M2}} = \SI{35.3}{\s}$ and $t_{\text{M3}} = \SI{38.6}{\s}$. 
The objective results are depicted in Fig.~\ref{plot:boxObjective}.

\subsection{Human experience} \label{sec:resultsSubjective}

\begin{figure}[tbp]
    \centering
    \resizebox{.48\textwidth}{!}{
        \input{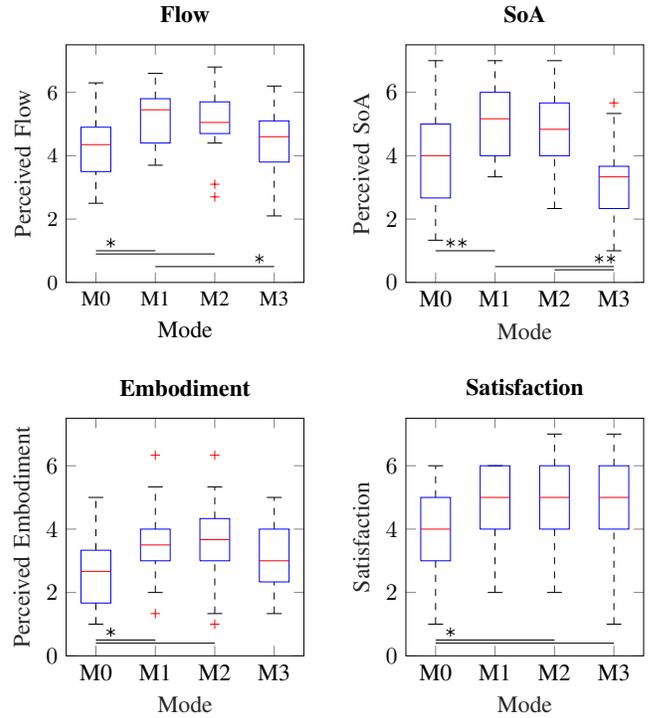}
        }
    \caption{Box plots showing human experience (\textit{flow}, \textit{SoA}, \textit{embodiment} and work result satisfaction) for different support levels (M0–M3). \textit{Flow}, \textit{SoA} and \textit{embodiment} scores are elevated in M1 and M2. Work result satisfaction plateaus with M1. }
    \label{plot:boxExperience}
\end{figure}

For the subjective analysis, we examined how the independent variable of the automation mode affects human experience and how different experience measures are associated with each other.

\subsubsection{Effect of automation level}

The influence of automation variation on the human experience was analyzed based on the psychological measures introduced in Section~\ref{subsubsec:humExp}. 

For the analysis of \textit{flow}, sphericity was assumed to be given ($Mauchly-W(5)=0.748$, $p =0.25$). 
The ANOVA showed that changes in the level of support led to significantly different mean \textit{flow} experiences ($F(3,75)=11.3$, $p<0.001$). 
Pairwise comparisons showed that M1 ($M=5.23$) and M2 ($M=5.06$) led to significantly higher mean \textit{flow} experiences than M0 ($M=4.25$) with $p< 0.014$. 

Further, M1 revealed significantly higher levels of \textit{flow} experience than M3 ($M=4.4$) with $p=0.011$, while the difference of M3 and M2 did not reach significance with $p=0.075$.

For \textit{SoA}, significant differences in experience were observed ($F(3,75)=11.4$ with a GG-corrected $p<0.001$ due to no sphericity with $Mauchly-W(5)=0.376$, $p=0.0003$). Here, M3 ($M=3.26$) showed a significantly lower mean \textit{SoA} score than M1 ($M=5.21$) and M2 ($M=4.69$) with $p\leq 0.001$. M0 ($M=3.99$) also resulted in a significantly lower mean \textit{SoA} than M1 ($p=0.0076$).

Regarding the influence of the automation mode on \textit{embodiment} measures, a significant effect was observed with $F(3,100)=4.4, p=0.006$ (sphericity was given with $Mauchly-W(5)=0.873$ and $p=0.667$). For this measure, M0 ($M=2.63$) resulted in a significantly lower mean score compared to M1 ($M=3.55$) and M2 ($M=3.62$) with $p<0.02$.

The effect of the automation on work result satisfaction was significant with $F(3,100) = 3.9, p=0.011$ (sphericity given with $Mauchly-W(5)=0.926$, $p=0.88$). 
M2 ($M=4.85$) and M3 ($M=4.85$) were rated significantly higher than M0 ($M0=3.77$).

\subsubsection{Correlation of subjective measures} \label{sec:resultsCorrSub}
In our analysis of subjective measures, we identified correlations between all three measures:  \textit{Flow} and \textit{SoA} showed a strong correlation ($r = 0.67$, $p<0.001$). 
Additionally, \textit{flow} and \textit{embodiment} ($r=0.46$, $p<0.001$) as well as \textit{SoA} and \textit{embodiment} ($r=0.41$, $p<0.001$) presented medium correlations.
 When averaging the answers to the four verbal questions per mode and per person, we observed a strong correlation with the full \textit{flow} questionnaire ($r=0.78$, $p<0.001$) and an even slightly stronger correlation with the smooth process subset of the \textit{flow} questionnaire ($r=0.80$, $p<0.001$).


\subsection{Grip force} \label{sec:resultsGrip}

\begin{figure}[tbp]
    \centering
    \resizebox{0.48\textwidth}{!}{
%
%
\begin{tikzpicture}

\begin{axis}[%
width=1.4in,
height=1.4in,
at={(0.5in,0.39in)},
scale only axis,
unbounded coords=jump,
xmin=0.5,
xmax=4.5,
xtick={1,2,3,4},
xticklabels={{M0},{M1},{M2},{M3}},
xlabel style={font=\color{white!15!black}},
xlabel={Mode},
ymin=0,
ymax=70,
ylabel style={font=\color{white!15!black}},
ylabel={Grip Force (N)},
axis background/.style={fill=white},
title style={font=\bfseries},
title={Mean Grip Force}
]
\addplot [color=black, dashed, forget plot]
  table[row sep=crcr]{%
1	44.0143075081644\\
1	60.277428515548\\
};
\addplot [color=black, dashed, forget plot]
  table[row sep=crcr]{%
2	35.9679786549102\\
2	54.35708144149\\
};
\addplot [color=black, dashed, forget plot]
  table[row sep=crcr]{%
3	33.4286765215572\\
3	53.7791174673543\\
};
\addplot [color=black, dashed, forget plot]
  table[row sep=crcr]{%
4	38.3653133847538\\
4	58.682206954607\\
};
\addplot [color=black, dashed, forget plot]
  table[row sep=crcr]{%
1	11.9901427283923\\
1	20.6529510814224\\
};
\addplot [color=black, dashed, forget plot]
  table[row sep=crcr]{%
2	11.274669844557\\
2	21.3177350232997\\
};
\addplot [color=black, dashed, forget plot]
  table[row sep=crcr]{%
3	8.62805933614322\\
3	18.1970780334655\\
};
\addplot [color=black, dashed, forget plot]
  table[row sep=crcr]{%
4	10.1210530822817\\
4	17.2150897515872\\
};
\addplot [color=black, forget plot]
  table[row sep=crcr]{%
0.875	60.277428515548\\
1.125	60.277428515548\\
};
\addplot [color=black, forget plot]
  table[row sep=crcr]{%
1.875	54.35708144149\\
2.125	54.35708144149\\
};
\addplot [color=black, forget plot]
  table[row sep=crcr]{%
2.875	53.7791174673543\\
3.125	53.7791174673543\\
};
\addplot [color=black, forget plot]
  table[row sep=crcr]{%
3.875	58.682206954607\\
4.125	58.682206954607\\
};
\addplot [color=black, forget plot]
  table[row sep=crcr]{%
0.875	11.9901427283923\\
1.125	11.9901427283923\\
};
\addplot [color=black, forget plot]
  table[row sep=crcr]{%
1.875	11.274669844557\\
2.125	11.274669844557\\
};
\addplot [color=black, forget plot]
  table[row sep=crcr]{%
2.875	8.62805933614322\\
3.125	8.62805933614322\\
};
\addplot [color=black, forget plot]
  table[row sep=crcr]{%
3.875	10.1210530822817\\
4.125	10.1210530822817\\
};
\addplot [color=blue, forget plot]
  table[row sep=crcr]{%
0.75	20.6529510814224\\
0.75	44.0143075081644\\
1.25	44.0143075081644\\
1.25	20.6529510814224\\
0.75	20.6529510814224\\
};
\addplot [color=blue, forget plot]
  table[row sep=crcr]{%
1.75	21.3177350232997\\
1.75	35.9679786549102\\
2.25	35.9679786549102\\
2.25	21.3177350232997\\
1.75	21.3177350232997\\
};
\addplot [color=blue, forget plot]
  table[row sep=crcr]{%
2.75	18.1970780334655\\
2.75	33.4286765215572\\
3.25	33.4286765215572\\
3.25	18.1970780334655\\
2.75	18.1970780334655\\
};
\addplot [color=blue, forget plot]
  table[row sep=crcr]{%
3.75	17.2150897515872\\
3.75	38.3653133847538\\
4.25	38.3653133847538\\
4.25	17.2150897515872\\
3.75	17.2150897515872\\
};
\addplot [color=red, forget plot]
  table[row sep=crcr]{%
0.75	30.3584118362443\\
1.25	30.3584118362443\\
};
\addplot [color=red, forget plot]
  table[row sep=crcr]{%
1.75	28.4359424983347\\
2.25	28.4359424983347\\
};
\addplot [color=red, forget plot]
  table[row sep=crcr]{%
2.75	26.8926048715533\\
3.25	26.8926048715533\\
};
\addplot [color=red, forget plot]
  table[row sep=crcr]{%
3.75	24.8053790438783\\
4.25	24.8053790438783\\
};
\addplot [color=black, only marks, mark=+, mark options={solid, draw=red}, forget plot]
  table[row sep=crcr]{%
nan	nan\\
};
\addplot [color=black, only marks, mark=+, mark options={solid, draw=red}, forget plot]
  table[row sep=crcr]{%
nan	nan\\
};
\addplot [color=black, only marks, mark=+, mark options={solid, draw=red}, forget plot]
  table[row sep=crcr]{%
nan	nan\\
};
\addplot [color=black, only marks, mark=+, mark options={solid, draw=red}, forget plot]
  table[row sep=crcr]{%
nan	nan\\
};
\end{axis}

\begin{axis}[%
width=1.4in,
height=1.4in,
at={(2.5in,0.39in)},
scale only axis,
unbounded coords=jump,
xmin=0.5,
xmax=4.5,
xtick={1,2,3,4},
xticklabels={{M0},{M1},{M2},{M3}},
xlabel style={font=\color{white!15!black}},
xlabel={Mode},
ymin=0,
ymax=4.5,
ylabel style={font=\color{white!15!black}},
ylabel={Grip Force Rank},
axis background/.style={fill=white},
title style={font=\bfseries},
title={Ranked Grip Force}
]
\addplot [color=black, dashed, forget plot]
  table[row sep=crcr]{%
1	4\\
1	4\\
};
\addplot [color=black, dashed, forget plot]
  table[row sep=crcr]{%
2	4\\
2	4\\
};
\addplot [color=black, dashed, forget plot]
  table[row sep=crcr]{%
3	3\\
3	4\\
};
\addplot [color=black, dashed, forget plot]
  table[row sep=crcr]{%
4	3\\
4	4\\
};
\addplot [color=black, dashed, forget plot]
  table[row sep=crcr]{%
1	1\\
1	2\\
};
\addplot [color=black, dashed, forget plot]
  table[row sep=crcr]{%
2	1\\
2	2\\
};
\addplot [color=black, dashed, forget plot]
  table[row sep=crcr]{%
3	1\\
3	1\\
};
\addplot [color=black, dashed, forget plot]
  table[row sep=crcr]{%
4	1\\
4	1\\
};
\addplot [color=black, forget plot]
  table[row sep=crcr]{%
0.875	4\\
1.125	4\\
};
\addplot [color=black, forget plot]
  table[row sep=crcr]{%
1.875	4\\
2.125	4\\
};
\addplot [color=black, forget plot]
  table[row sep=crcr]{%
2.875	4\\
3.125	4\\
};
\addplot [color=black, forget plot]
  table[row sep=crcr]{%
3.875	4\\
4.125	4\\
};
\addplot [color=black, forget plot]
  table[row sep=crcr]{%
0.875	1\\
1.125	1\\
};
\addplot [color=black, forget plot]
  table[row sep=crcr]{%
1.875	1\\
2.125	1\\
};
\addplot [color=black, forget plot]
  table[row sep=crcr]{%
2.875	1\\
3.125	1\\
};
\addplot [color=black, forget plot]
  table[row sep=crcr]{%
3.875	1\\
4.125	1\\
};
\addplot [color=blue, forget plot]
  table[row sep=crcr]{%
0.75	2\\
0.75	4\\
1.25	4\\
1.25	2\\
0.75	2\\
};
\addplot [color=blue, forget plot]
  table[row sep=crcr]{%
1.75	2\\
1.75	4\\
2.25	4\\
2.25	2\\
1.75	2\\
};
\addplot [color=blue, forget plot]
  table[row sep=crcr]{%
2.75	1\\
2.75	3\\
3.25	3\\
3.25	1\\
2.75	1\\
};
\addplot [color=blue, forget plot]
  table[row sep=crcr]{%
3.75	1\\
3.75	3\\
4.25	3\\
4.25	1\\
3.75	1\\
};
\addplot [color=red, forget plot]
  table[row sep=crcr]{%
0.75	3\\
1.25	3\\
};
\addplot [color=red, forget plot]
  table[row sep=crcr]{%
1.75	3\\
2.25	3\\
};
\addplot [color=red, forget plot]
  table[row sep=crcr]{%
2.75	2\\
3.25	2\\
};
\addplot [color=red, forget plot]
  table[row sep=crcr]{%
3.75	2\\
4.25	2\\
};
\addplot [color=black, only marks, mark=+, mark options={solid, draw=red}, forget plot]
  table[row sep=crcr]{%
nan	nan\\
};
\addplot [color=black, only marks, mark=+, mark options={solid, draw=red}, forget plot]
  table[row sep=crcr]{%
nan	nan\\
};
\addplot [color=black, only marks, mark=+, mark options={solid, draw=red}, forget plot]
  table[row sep=crcr]{%
nan	nan\\
};
\addplot [color=black, only marks, mark=+, mark options={solid, draw=red}, forget plot]
  table[row sep=crcr]{%
nan	nan\\
};
\end{axis}
\end{tikzpicture}%
        }
    \caption{Grip force decreases for an increase in automation support.  }
    \label{plot:gripforce}
\end{figure}
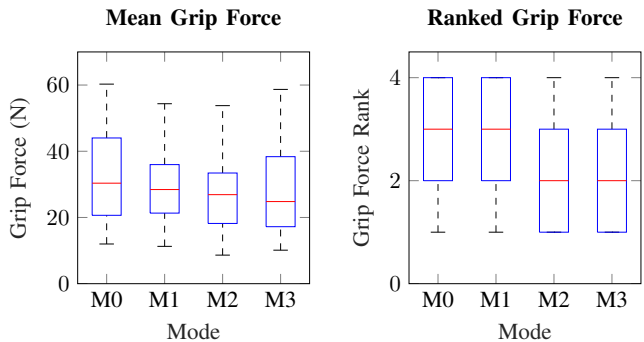

The grip force exerted on the cordless drill handle by each subject was recorded for the duration of the interaction. The mean force over all boards per mode was calculated and depicted in Fig.~\ref{plot:gripforce}.
Additionally, we sorted the modes per subject by decreasing mean grip force: Using a Spearman ranked correlation test, we observed a medium-strong correlation between the level of automation and ranked grip force ($r = -0.30$,  $p=0.0026$). 
When examining the correlations between grip force and experience measures, we identified a weak correlation between ranked grip force and \textit{SoA} ($r=0.20$, $p=0.045$).

\subsection{Heart-rate variability}

Our goal was to predict participants’ \textit{flow} values, obtained from a questionnaire, using HRV data. This investigation aimed to determine whether sensor-based measurements can serve as a viable alternative to self-reported questionnaire data. Due to the limited amount of data, we relied on handcrafted feature extraction to capture relevant patterns in the HRV signals.

We analyzed HRV data following the methodology presented in Rissler et al.~\cite{Rissler.2023}. After removing erroneous measurements caused by connectivity issues using a median filter, we extracted both time-domain and frequency-domain features. Feature computation was performed using the Python package \texttt{pyhrv} \cite{pyhrv}.

In the time domain, we calculated the mean and standard deviation of heart rate, the mean and standard deviation of RR intervals, the root mean squared of successive differences (RMSSD) of RR intervals, and the percentage of RR successive differences exceeding \SI{50}{\ms}. In the frequency domain, we computed the energy associated with the very low frequency (VLF), low frequency (LF), and high frequency (HF) components, the total power of the spectral density, the LF and HF power normalized by total power, and the LF/HF power ratio.

Following Rissler et al.~\cite{Rissler.2023}, we excluded data where the difference between a participant’s \textit{flow} score and the median \textit{flow} score is below 0.3. This filtering step resulted in 65 valid measurements with binary class labels.

For prediction, we employed a random forest model with 50 trees, a maximum depth of 7, and a minimum of 10 samples per leaf node. Given the limited dataset size, we applied cross-validation with a 60-to-5 split. We report a mean accuracy of 60.95\%, averaged over 50 different random seeds, and 13 cross-validation test sets.

\section{Discussion}  \label{sec:discussion}

In this section, we discuss the previously presented results and derive their implications.

\subsection{Control accuracy}

As illustrated in Section~\ref{sec:resultsObjective}, the automation design impacted the task performance in the expected manner - as the level of automation increases, the joint task was performed with less errors. 
The increase in accuracy is particularly evident in the rotational specification concerning the $B$- and $C$-axes, as well as the variance in height. 
With regard to the $y$-equidistance, a modest inclination towards reduced error is observed as support level rises. 
The results indicate that humans have difficulty maintaining a given rotational drill angle but demonstrate proficiency in tracing set positions within the translational domain. Overall, an enhancement in objective performance is evident with each increment of support.

\subsection{Human experience}

Results presented in \ref{sec:resultsSubjective} revealed that automation level significantly influences human experience. According to flow theory, engagement is highest when task difficulty matches user skill, and both insufficient and excessive automation can lead to disengagement or frustration~\cite{Csikszentmihalyi.1990, Csikszentmihalyi.1989}. Our findings support this hypothesis, showing that moderate automation levels (M1, M2) yield the highest flow, SoA, and embodiment, while too little (M0) or too much (M3) automation reduces subjective experience. 
The level of automation and user engagement and experience reveal an inverted-U shape, where both extremes in automation diminish user engagement and experience~\cite{Parasuraman.2000}. These results align with adaptive automation models, which emphasize dynamic adjustments to automation levels to maintain cognitive involvement and prevent both overload and passive disengagement~\cite{Kaber.2004a}. This finding is particularly noteworthy for two reasons. First, despite users having full manual control in M0, their SoA and embodiment ratings are higher in M1 and M2. This seemingly counterintuitive result aligns with motor prediction models, which argue that SoA is not simply derived from manual effort but rather from the predictability and alignment of action outcomes with user intentions~\cite{Blakemore.2002, Wen.2015}. Moderate automation enhances SoA by providing predictable, responsive assistance that reinforces the user’s intended actions rather than overriding them~\cite{Abbink.2012}. This suggests that even with some level of external support, users feel a stronger sense of responsibility for their actions compared to fully manual conditions, supporting the notion that shared control can enhance, rather than diminish, agency~\cite{Damen.2015}. Second, although task errors decrease in M3, the perceived flow score is lower, likely due to restricted user action. Flow emerges when an individual is fully engaged in a task that presents an optimal balance between challenge and skill. Therefore, excessive automation disrupts this balance, reducing intrinsic motivation and engagement by limiting the user’s meaningful participation in the task. This hypothesis is further supported by the observed correlation between SoA and flow, which suggests that feeling in control of one’s actions is a prerequisite for deep task engagement~\cite{Haggard.2012}. 
 
 The subjective satisfaction with task outcomes is similar between M2 and M3, indicating that users do not perceive the slight increase in accuracy from M2 to M3 as meaningful for their overall experience. This suggests that beyond a certain threshold, further automation precision does not proportionally improve user satisfaction, potentially transferring the hypothesis of diminishing returns~\cite{Shephard.1974} to automation design. Even further, increased automation can instead lead to complacency, reduced situational awareness, and disengagement~\cite{Parasuraman.2000a, Parker.2022}. While the highest level of automation support results in optimal objective performance, a moderate level of automation leads to higher subjective experience scores, reflecting a balance between system guidance and user agency. Furthermore, strong correlations between flow, SoA, and embodiment suggest that these constructs are interconnected in human-robot interaction, aligning with research indicating that flow, SoA, and embodiment share common cognitive and affective mechanisms, particularly in physically interactive tasks where sensorimotor integration plays a key role~\cite{Kilteni.2012}. Taken together, these findings suggest that an intermediate level of automation optimizes user experience across flow, SoA, and embodiment. Future research should explore whether these experience measures can be independently modulated.

\subsection{Grip force}

We observe a medium negative correlation ($r=-0.3$) between the level of support and the level of grip force, showing that the participants loosen their grip when the automation increases its dominance. This suggests that the human reduces their control by reducing their grip force, allowing the automation to contribute more to the task. This aligns with the observation of a positive correlation between the level of grip force and \textit{SoA} ($r=0.2$): The more the human feels responsible for an action, the stronger the grip force. 
The fact that the correlation exists only for the subject-specific ranked grip force and not the absolute grip force, indicates that it is a highly subject-dependent measure and a baseline measurement is advised.

\subsection{Heart-rate variability}

We aimed to replicate the promising findings of Rissler et al.~\cite{Rissler.2023}. Our hypothesis was that \textit{flow} experience would correlate with specific HRV features. However, the results of our study did not reinforce their findings. 
With a mean accuracy of only 60.95\%, the predictive performance remains limited. This limitation may be due to an insufficient number of data points for feature extraction and a trial duration too short to capture physiological \textit{flow}.

\subsection{Limitations}
While this study provides valuable insights into the relationship between automation design and human experience in human-robot collaboration, several limitations must be acknowledged.

First, the generalizability of our findings to other application domains remains uncertain. Our experimental task — a collaborative wood workshop scenario — was designed to simulate real-world human-robot interaction, yet it may not fully capture the complexities of other domains, such as industrial assembly, surgical robotics, or teleoperation. Future research should examine whether similar effects on flow, SoA and embodiment emerge in different contexts with varying task demands and environmental constraints.

Second, we did not explicitly assess how strongly participants perceived the provided automation. While our results suggest that intermediate levels of automation are most beneficial for human experience, the absence of direct measurements on perceived support intensity limits our ability to draw definitive conclusions about individual variability in automation perception. Future studies should incorporate subjective ratings or psychophysical methods to quantify perceived assistance levels.

Third, individual differences in physical abilities may have influenced our findings. While we observe correlations between grip force and SoA, we did not control for baseline grip strength or participants' habitual grip force levels. Variability in physical ability may have led to different interaction strategies, potentially affecting both the physiological and subjective experience measures. Incorporating personalized calibration or normalizing grip force data relative to each participant’s baseline could enhance the robustness of future analyses.

Additionally, our study primarily involved participants with no prior professional experience in the given task. The extent to which professional users — such as experienced craftworkers or robotics specialists — would respond similarly remains unknown. Skilled users might exhibit different adaptation behaviors, expectations, and sensitivities to automation support. Future work should investigate whether expertise modulates the effects of automation design on human experience.

A limitation of our flow analysis is the relatively short interaction duration, which may have been insufficient for participants to fully experience a deep flow state. Flow typically emerges over extended periods of engagement, allowing for progressive immersion and adaptation to the task. Given the time constraints of our study, the measured flow levels may not reflect the full extent of the phenomenon, potentially explaining the inconclusive relationship between flow and HRV. Future studies should explore longer task durations to assess whether prolonged exposure to automation influences physiological correlates of flow more consistently.

Finally, the inertia of the robotic system may have influenced participants' perception of the interaction. The physical dynamics of the robotic system, including its mass and damping properties, might have introduced an unintended resistance that shaped the subjective experience. However, since we did not compare our setup against a fully manual condition, we cannot determine whether this effect was a limiting factor or whether a similar trend would emerge in other robotic implementations. Future research should systematically compare robotic and manual conditions to isolate the impact of robot dynamics on human experience.

Despite these limitations, our findings provide a crucial step toward understanding how automation design influences human experience in collaborative robotics. Addressing these limitations in future studies will further refine our understanding and support the development of more intuitive, human-centric automation strategies.

\section{Conclusion}  \label{sec:conclusion}

This work proposes a collaboration task and study design that allows for the analysis of the effect of automation on human experience measures such as flow, SoA and embodiment. 
A user study was conducted with 28 participants, each performing a collaborative woodworking task using four different levels of automation and collecting multiple objective and subjective measures. 
The subsequent analysis of the interaction yielded several findings: The level of automation exhibits a significant influence on the experience measures under investigation. Specifically, a medium level of support ranks highest in flow, SoA, and embodiment, even though the strongest level of support results in enhanced task performance. Additionally, an increase in task performance after a certain threshold does not lead to an increase in human's work result satisfaction, transferring the diminishing returns hypothesis to performance in human-robot interaction. 
These findings all underscore the necessity for an automation optimization that encompasses human experience measures next to task performance. 
    
While the replication of other findings that suggest using HRV features as an indicator for flow experience was unsuccessful, grip force was identified as a potential measure of SoA, which could substitute or complement experience assessments using questionnaires. It could therefore serve as a measure to adapt the level of automation, potentially increasing the human experience in interaction. 
The integration of these findings contributes to the advancement of human-robot systems, paving the way for a symbiotic relationship between humans and machines.

\section*{Appendix}

\subsection{Questionnaires}

\underline{\textit{Flow}} (7-pts Likert-scale)
\begin{itemize}
    \item Unadapted Flow Short Scale (FSS) by~\cite{Rheinberg.2003}
\end{itemize}

\underline{\textit{Sense of Agency (SoA})} (7-pts Likert-scale)

Adapted from~\cite{Tapal.2017}:
\begin{itemize}
    \item I can move the machine exactly the way I want, as if the machine was obeying my will. (Ich kann die Maschine genau so bewegen wie ich es möchte, als würde die Maschine meinem Willen gehorchen.)
    \item I have the feeling that I can control the movements of the machine. (Ich habe das Gefühl die Bewegungen der Maschine zu kontrollieren.)
    \item The movements of the machine were caused by my actions. (Es fühlte sich so an, als würde ich die Bewegungen, die ich bei der Maschine gesehen habe, selbst verursachen.)
\end{itemize}

\underline{\textit{Embodiment}} (7-pts Likert-scale)

Adapted from~\cite{Gonzalez-Franco.2018}: 
\begin{itemize}
    \item I can operate the machine as easily as
I can use my arms and legs. (Ich kann die Maschine so einfach bedienen wie ich meine Arme und Beine benutzen kann.)
\item It feels as if the machine is part of my body. (Es fühlt sich so an, als wäre die Maschine Teil meines Körpers.)
\item It felt as if I could feel the movements that the machine was making. (Es fühlte sich so an, als könnte ich die Bewegungen, die die Maschine ausführte, spüren.)
\end{itemize}

\underline{Work result satisfaction} (7-pts Likert-scale)
\begin{itemize}
    \item I am satisfied with the result of my work. (Ich bin mit dem Ergebnis meiner Arbeit zufrieden.)
\end{itemize}

\addtolength{\textheight}{-4cm}   

\bibliography{1_tex/files/bibliography}
\bibliographystyle{IEEEtran}

\end{document}